# Malicious Code Execution Detection and Response Immune System inpired by the Danger Theory

Jungwon Kim[†] , Julie Greensmith, Jamie Twycross and Uwe Aickelin
[†] Department of Computer Science, University College London, UK
j.kim@cs.ucl.ac.uk
School of Computer Science, University of Nottingham, UK
{jqg, jpt, uxa}@cs.nott.ac.uk

**Abstract-** **The analysis of system calls is one method em- ployed by anomaly detection systems to recognise mali- cious code execution. Similarities can be drawn between this process and the behaviour of certain cells belong- ing to the human immune system, and can be applied to construct an artificial immune system. A recently developed hypothesis in immunology, the Danger Theory, states that our immune system responds to the presence of intruders through sensing molecules belonging to those invaders, plus signals generated by the host indic- ating danger and damage. We propose the incorporation of this concept into a responsive intrusion detection sys- tem, where behavioural information of the system and running processes is combined with information regard- ing individual system calls.**

## 1 Introduction

Malicious code execution through the exploitation of software vulnerabilities can allow an intruder to compromise a host running the software. The running of any process on a machine generates system calls, providing interaction between application, operating system and hardware. Analysis of the system calls made by a process, through the verification of system call usage, can reveal the execution of malicious code. Previous research attempted to identify the exploitation vulnerable software by detecting anomalies present in system call ('syscall') traces [hofm98][krug03]. Research in this area frequently utilises the construction of system call profiles during the legitimate operation of monitored programs. During the detection process, any syscall sequences or arguments that do not comply with the previously generated 'normal' profiles are regarded as a sign that the system is compromised. Although these approaches have produced promising results, they can produce high rates of false positive errors, an issue which has yet to be resolved. As outlined in [tan02], this may arise from the fact that the environment comprising the 'system' is largely ignored

The key objective of our work is to propose a solution to this problem by taking inspiration from the Human Immune System (HIS). A hotly debated hypothesis in immunology, known as the Danger Theory [matz94][matz02] illustrates that the HIS can detect danger in addition to the collection of proteins known as antigens in order to trigger appropri- ate immune responses. Similarly, in the case of utilising sy- scalls to detect malicious code execution, we argue that the presentation of data must be coupled with its environmental conditions to create a sense of context.

Previous research into computer security within the biologically inspired field of Artificial Immune Systems (AIS) has focused on detecting intrusions through the use of algorithms based on the self-nonself theory of immunogenicity. This theory dictates that the immune system is activated by the presence of 'non-self' or foreign proteins, and is a prevalent hypothesis in immunology[aick04]. However, there are numerous instances where this classification fails. For instance, there is no immune response to foreign bacteria in the gut or to food and conversely, some autoreactive processes exist, causing diseases such as rheumatoid arthritis and multiple sclerosis. The Danger Theory challenges this viewpoint, arguing that foreign invaders cause damage to the host, thus inducing the release of cell molecules (termed danger signals) by detecting cell stress and death[matz94]. These signals are exposed to antigen presenting cells, which in turn initiate immune responses, due to the correlation of the signals. Similarly, we propose a system which encompasses various features of the danger theory to provide context to existing system call data, and generating an appropriate response.

## 2 Approach

### 2.1 Dendritic Cells and Danger Signals

Dendritic cells (DCs) are 'professional' antigen presenting cells, specialised for presenting collected proteins (antigen) in combination with their environmental context. This information is presented to effector T cells [manf02], causing the recognition and removal of pathogens. Following migration to the lymph node, antigen is displayed with context signals by DCs, which can activate 'naive' T cells. T cells expressing a complimentary receptor for the antigen are activated if presented in a dangerous or 'necrotic' context. Conversely, a safe or 'apoptotic' context causes any matching T cells become tolerised to that particular protein.

The context information is translated using the differentiation pathways of the DCs. DCs exist in three maturation states: immature(iDC), semi-mature (smDC) and mature (mDC). Initially, when a DC enters the tissue, it exists in an immature state. The function of an iDC is to collect cellular debris from the tissue via ingestion. Debris comprising of protein is extracted and stored, in preparation for presentation. However, presentation of antigen to T cells via iDCs results in the deactivation of the T cell, as it does not express the necessary costimulatory molecules or inflammatory cytokines (local chemical messengers) required for full T cell activation.

However, if the tissue is damaged as a result of a patho-genic infection or other cell stress, 'danger signals' are released into the tissue. Additionally, pathogens like bacteria express proteins which can be recognised through specific receptors (pattern recognition receptors) on DCs, which have evolved over millions of years and are highly conserved. These pathogenic proteins are known as pathogen associated molecular patterns (PAMPs). Exposure to PAMPS, danger signals or both, causes the full maturation of an iDC, initiating migration out of the tissue to the lymph node. On disruption of the membrane, a cell undergoes lysis and releases all of its contents into the surrounding area. A number of molecules found only inside of cells, such as uric acid, appear in the interstitial fluid. The signals are an indicator of cell stress, implying danger is present within that particular tissue. Inflammatory cytokines produced by other mDCs in the area can have an amplicative effect on both the PAMPs and danger signals. The cellular effects of exposure to PAMPS and danger signals result in increased production of the costimulatory molecules necessary for T cell binding, and the expression of cytokines which activate naive T cells. This can lead to a full adaptive immune response.

Conversely, if the tissue is healthy and the cells are not under stress, apoptosis is the dominant kind of cell death, resulting in the regulated dismantling of the cell. This ensures that the cell contents are disposed without entering the interstitial fluid. The presence of cytokines released as a result of apoptosis, bind to different receptors on the DC, again, modifying the output cytokines expressed. This results in the increased production of the costimulatory molecules (as with the mDC), but the increased production of different cytokines. The cytokines released by the so called 'semi-mature' DCs are thought to tolerise T-cells to the antigen presented, and to produce regulatory T-cells which also have a suppressive effect.

Essentially, DCs have the capacity to act as biological anomaly detectors [gree05]. They combine multiple signal inputs (in the form of PAMPs, danger signals, inflammatory cytokines and apoptotic signals), process volumes of antigen and provide T cells with essential context information regarding the health of the tissue. An abstracted model of these cells forms the basis of the anomaly detection component of our system, given their ability as natural danger detectors. In order to apply a dendritic cell inspired algorithm for the detection of malicious code execution, a suitable mapping for antigen and signals must be formulated.

Each DC is programmed with a simple multi-signal processor and antigen collector function. The signals represent the context for the collected antigen[1], and a unique signal mapping schema is applied. PAMPS are represented by violations of security policies created by a system call checker such as systrace [prov03]. These are signature based signals, like proteins that we know to come from pathogenic sources. The danger signals represent the behaviour of the process, with deviations in the behaviour increasing the level of danger within the system. Conversely, the continuous normal behaviour of the processes generates safe signals. Inflammatory cytokines are not enough to initiate full DC maturation, but give an indication of the health of the tissue, and therefore can be mapped to examining the general behaviour of the host machine. Specific examples and a summary of the four types of signal are provided in Table 1.

| Signals | Meaning | Examples |
|---|---|---|
| PAMP | known to be pathogenic | Security policy violation |
| Safe Signals | indicates stable/normal conditions | No detectable change in process cpu load or memory usage |
| Danger Signals | may indicate changes in behaviour | Highly fluctuating process cpu or memory usage |
| Inflammatory Cytokines | amplify the effects of the other signals | System load average |

Table 1: Signals provided for malicious process detection

The data comprising the 'antigen' and 'signals' are captured and stored ready for collection by a subset of the DC population. The signals are received and stored for collection at various concentrations, relative to the level of deviation. Virtual output cytokines derived through combining the inputs are used to direct our response-generating T cells. The virtual DCs have three types of output cytokine concentrations: costimulatory molecules (CSM), mature cytokines and semi-mature cytokines. High concentrations of PAMPs and danger signals lead to a high expression of mature cytokines, which can be further amplified in value by the addition of inflammatory cytokines. Conversely, the apoptotic signals are used to increase semi-mature cytokine expression, and to suppress the effects of PAMPs. The receipt of any kind of signal results in an increased level of CSM expression. Once the concentration of CSM reaches a threshold, the DC can collect no more antigen or signals, and is removed from the tissue and sent to the lymph node.

2.2 Antigens and T cell Maturation

In an immunological context, antigens are defined as substances which can initiate adaptive immune responses [coic03]. When applied to our system, the execution of a program - a process, can be regarded as a collection of antigens. Should the program be exploited, this will change the expected behaviour of the process. Once this is detected by the DC population, the T cell component of the system must respond accordingly - delaying or stopping the execution of the malicious code. Therefore, system calls comprising the process being exploited can be viewed as antigens, and have the capacity to stimulate an active response. As a running process generates a large number of system calls, antigens are a subset of the total syscalls executed.

Our system captures antigens (sets of syscalls) by using a system call policy checker tool, systrace [syst03] .

---

[1] The details of an antigen used by the AIS are discussed in section 2.2.

These antigens are further fragmented by the DC population, forming antigen-peptides, a further subset of the total syscalls for a scrutinised executable. Various sets of antigen peptide can be extracted from an antigen multiple times, by multiple DC. This is both in terms of different sequence lengths and sampling rates. Hence, a population of DCs present a set of system calls in diverse forms together with antigen contexts, which are represented by output cytokine concentrations.

In the HIS, DCs interact with naive T cells which have the capacity to differentiate further into effector T cells [coic03]. Naive T cells are newly created cells which have not encountered antigen and do not yet exhibit immune response functions, such as cell destroying abilities. They have a specialised receptors type called TCRs which vary greatly between T cells. This receptor facilitates the binding between the TCR and presented antigen peptides. Naive T cells are activated when the TCR- antigen peptide binding affinity is sufficiently high and the necessary CSMs and cytokines are present in large enough concentrations. The activated T cells gain the ability to proliferate and their clones begin to differentiate into effector T cells. These cells present in distressed or inflamed tissues recognise infected cells by binding their TCRs to pathogenic antigen peptides on target cell surfaces.

In the AIS, naive T cells are generated when smDCs or mDCs carry antigen peptides to artificial lymph nodes. TCRs of naive T cells are created by randomly sampling, combining and generalising the antigen peptides passed from the DCs. Specifically, different combinations of syscall sequences and partial strings of syscall arguments become TCRs of naive T cells, which can be signatures of an on-going attack. Our AIS uses an individual policy statement, which is generated by systrace, to represent TCRs of naive T cells. Systrace itself monitors system calls produced by a running program and it forces actions such as denying or permitting system calls based on a pre-defined policy [prov03]. The system call policy consists of a number of policy statements, which have a 'condition' part describing the states of monitored system calls and an 'action' part addressing actions to be taken when the condition part is observed. We define the condition part of the individual policy statement as the TCR, and the action portion as various types of computer immune responses.

In order for naive T cells to acquire activation, they constantly interact with smDCs and mDCs present in the lymph nodes. Whenever DCs interact with naive T cells, the presented antigen peptides are evaluated based on satisfactory matching with the TCRs (the condition parts of individual policy statements) of the selected T cell. Following a successful match, the three output cytokine concentrations of the presenting DC are examined. Naive T cells interpret three categories of DC output cytokines. High concentrations of mDC cytokines are translated attack associated antigens, with high concentrations of semi-mature cytokines indicating normal system function. Additionally, the value of expressed CSM indicates that the DC has been exposed to sufficient signals to permit presentation. Hence, if single antigens are captured at regular time intervals, the CSM value indicates that the output cytokines are accumulated over a long period time. This kind of information can provide more fine-grained evidence to support the decisions made over the generation of the appropriate response, perhaps giving rise to fewer false positives. Naive T cells have two numerical values called activation values and tolerance values that reflect the accumulated results of DC output cytokine concentration examination. These values indicate the state of naive T cell maturation. Upon this value reaching a pre-defined threshold, naive T cells finally become effector T cells. The activation and tolerance values are increased by the mDC and smDC cytokine concentrations respectively.

2.3 T cell Differentiation and Responses

Whenever naive T cells are created, they are assigned a lifespan and an age. The lifespan indicates the maximum period of timesteps which the naive T cells can survive and interact with DCs, and the age represents the length of the T cell-DC interaction. When the activation or tolerance values of the naive T cells exceed the predefined thresholds before their ages reach the lifespans, the naive T cells become effector T cells. If the threshold is not reached the T cells are subject to permanent deletion. This acts to safeguard against a number of false positive and negative errors, that could arise as a result of over-generalisation on behalf of the T cell.

The type of response generated by the effector T-cell is dependent on which value reaches the threshold first. The allotted responses of the effector T cells are presented by the 'action' parts of the system call policy statements, previously used to represent the TCRs. If the activation value triggers T-cell differentiation, a 'deny' action is given as the response of the effector T cell. A 'permit' response is generated if differentiation does not occur. The effector T cells with TCRs and associated responses start monitoring new system call traces. Whenever new system calls satisfy the TCRs of effector T cells, the AIS takes the actions which are presented as the responses of the effector T cells. Two types of responses can be generated - system call permission and denial, both implemented using systrace.

In order to maintain a sense of homeostasis, effector T-cells have a controlled lifespan in addition to naive T-cells. New system call policies presented by newly differentiated effector T cells are applied only during their restricted lifespan. However, a subset of these effector T cells, appropriately termed memory T cells, can remain for a longer period of time than the initially defined lifespan. In the HIS, the approximately 90 percent of effector cells die after their responses or lifespans, the rest remain as memory [coic03]. Memory T cells are known to have an increased capacity for survival. Many factors contribute to the generation of 'effector memory', with the exact biological mechanism still to be determined. For our AIS, we are also currently examining a number of possible immunological mechanisms which explains the memory T cell pool maintenance, and evaluating whether any of these explanations might provide a way to maintain memory effector T cells.

## 3 Conclusion

This paper introduces a novel artificial immune system that detects malicious code execution and responds appropriately. The illustrated system implements various immunological mechanisms that are principally explained by the Danger Theory. The key assumption that our system attempts to verify is that our AIS is able to i) detect a danger from environmental conditions, ii) extracts and generalises attack signatures from the data associated with the detected danger, and iii) hence responds to an on-going attack appropriately.

The implementation of such a system has been progressing using systrace, which is a system call policy checker. Various signals describing the behaviours of monitored processes and hosts are processed by a population of dendritic cells (DCs). The processed signals coupled with antigens (sets of system calls) are presented to naive T cells. Three groups of signal processing outcomes lead naive T cells to differentiate to effector T cells, presenting attack signatures and a selected response, to permit or deny a system call.

Whilst the proposed system automatically updates the system call policy of systrace as a result of artificial immune responses, it does not aim to oppose the manual policy generation method currently used by systrace. Instead, our system is designed to extend an initial policy which is constructed defined by a user. There are already available publically-audited policies, including the Hairy Eyeball project. This project introduces systrace policies of nearly two hundred vulnerable programs [hair]. It is very common for a user to re-use these policies in order to save their time or select relatively reliable choices of policies. However, these policies do not cover all malicious or benign system calls regardless of whether they are manually predefined by a user, or reconfigured from publically-audited policies.

Currently systrace adopts two modes to handle system calls which do not match any policy statements. The user is asked to decide whether to deny or permit a system call. During this process the monitored program pauses until a decision is received, with the appropriate action then taken. In contrast, the artificial immune responses initiated by our system are expected to handle this situation automatically. The long hours of user absence or default denial can cause unnecessary program halt. In addition, the adopting publically-audited policies requires some reconfiguration of the policies to ensure suitability in the new environment. Artificial immune responses controlled by our system would be able to automate the reconfiguration of publically-audited policies. Similarly, our system would allow an automated update of policies in line with changes in the local computing environment.

## Acknowledgments

This project is supported by the EPSRC (GR/S47809/01), Hewlett-Packard Labs, Bristol, and the Firestorm intrusion detection system team. Great thanks to all the members of the "Danger Project" (www.dangertheory.com) for their helpful feedback and inspirational discussion.